\documentclass[twocolumn]{article}
\usepackage[utf8]{inputenc}

\usepackage{graphicx}

\usepackage{colortbl}
\usepackage[numbers]{natbib}
\usepackage{comment}
\usepackage{balance}
\usepackage{enumitem}
\usepackage{makecell}
\usepackage{array}
\usepackage{dirtytalk}

\usepackage{fancyhdr}
\usepackage{ragged2e}
\usepackage[final]{microtype}

\newcolumntype{L}[1]{>{\raggedright\let\newline\\\arraybackslash\hspace{0pt}}m{#1}}
\newcolumntype{C}[1]{>{\centering\let\newline\\\arraybackslash\hspace{0pt}}m{#1}}
\newcolumntype{R}[1]{>{\raggedleft\let\newline\\\arraybackslash\hspace{0pt}}m{#1}}

\fancypagestyle{firstpage}
{
    \fancyhf{}

    \fancyhead[L]{\small \justifying \noindent \textbf{Cite as: Bignold, A., Cruz, F., Taylor, M., Brys, T., Dazeley, R., Vamplew, P., Foale, C. (2021). A Conceptual Framework for Externally-influenced Agents: An Assisted Reinforcement Learning Review. Journal of Ambient Intelligence and Humanized Computing.} 
    }

    \fancyfoot[L]{\small \justifying \noindent This preprint has not undergone peer review or any post-submission improvements or corrections. The Version of Record of this article is published in the Journal of Ambient Intelligence and Humanized Computing (JAIHC), and is available online at https://doi.org/10.1007/s12652-021-03489-y.
    }
    
}

\title{A Conceptual Framework for Externally-influenced Agents:\\An Assisted Reinforcement Learning Review}
\author{
Adam Bignold$^{1,*}$ \and
Francisco Cruz$^{2,3,*}$ \and
Matthew E.~Taylor$^{4}$ \and
Tim Brys$^{5}$ \and
Richard Dazeley$^{2}$ \and
Peter Vamplew$^{1}$ \and
Cameron Foale$^{1}$
}
\date{\normalsize
$^1$ School of Engineering, IT and Physical Sciences, Federation University, Ballarat, Australia.\\
$^2$ School of Information Technology, Deakin University, Geelong, Australia.\\
$^3$ Escuela de Ingenier\'ia, Universidad Central de Chile, Santiago, Chile.\\
$^4$ Department of Computing Science and The Alberta Machine Intelligence Institute (Amii), University of Alberta, Edmonton, AB, Canada.\\
$^5$ Action Research Associates, Beirut, Lebanon.\\
$^*$ Both authors contributed equally to this manuscript.\\
Corresponding e-mails: \{a.bignold, p.vamplew, c.foale\}@federation.edu.au, \\ \{francisco.cruz, richard.dazeley\}@deakin.edu.au, matthew.e.taylor@ualberta.ca, tbrys@actionresearchassociates.org \\
}

\begin{document}

\maketitle

\begin{abstract}
A long-term goal of reinforcement learning agents is to be able to perform tasks in complex real-world scenarios. 
The use of external information is one way of scaling agents to more complex problems. 
However, there is a general lack of collaboration or interoperability between different approaches using external information. 
In this work, while reviewing externally-influenced methods, we propose a conceptual framework and taxonomy for assisted reinforcement learning, aimed at fostering collaboration by classifying and comparing various methods that use external information in the learning process. 
The proposed taxonomy details the relationship between the external information source and the learner agent, highlighting the process of information decomposition, structure, retention, and how it can be used to influence agent learning.
As well as reviewing state-of-the-art methods, we identify current streams of reinforcement learning that use external information in order to improve the agent's performance and its decision-making process. 
These include heuristic reinforcement learning, interactive reinforcement learning, learning from demonstration, transfer learning, and learning from multiple sources, among others. 
These streams of reinforcement learning operate with the shared objective of scaffolding the learner agent.
Lastly, we discuss further possibilities for future work in the field of assisted reinforcement learning systems.
\end{abstract}


\thispagestyle{firstpage}

\section{Introduction}
Reinforcement learning (RL)~\citep{sutton2018reinforcement} is a learning approach in which an agent uses sequential decisions to interact with its environment trying to find a (near-) optimal policy to perform an intended task.
RL agents have the ability to improve while operating, to learn without supervision, and to adapt to changing circumstances~\citep{kaelbling1996reinforcement}. 
By exploring, a standard agent learns solely from the signals it receives from the environment.
The RL approach has shown success in domains such as robotics~\citep{kitano1997robocup,  kober2013reinforcement, cruz2018action, contreras2020unmanned}, game-playing~\citep{tesauro1994td, barros2020moody}, inventory management~\citep{giannoccaro2002inventory}, and cloud computing~\citep{shakarami2020survey, shahidinejad2020joint, ghobaei2018learning}, among others.

Like many machine learning techniques, RL faces the problem of high-dimensionality spaces. 
As environments become larger, the agent's learning time increases and finding the optimal solution becomes impractical~\citep{cassandra2016learning}. 
Early research on this topic~\citep{kaelbling1996reinforcement, lin1991programming} argued that for RL to successfully scale into real-world scenarios, then the use of information external to the environment would be needed.
Different RL strategies using this approach have emerged in order to speed up the learning process.
They use external information to assist either the process of generalising the environment representation~\citep{price2003accelerating}, the agent's decision-making process~\citep{griffith2013policy}, or in providing more focused exploration~\citep{fernandez2006probabilistic}. 

In this article, we refer to \textit{external information} as any kind of information provided to the agent originating from outside of the agent's representation of the environment. 
This may include demonstrations~\citep{konidaris2012robot, rozo2013robot, chen2019active}, advice and critiques~\citep{knox2010combining,griffith2013policy}, initial bias based on previously gathered data~\citep{taylor2009transfer}, or highly-detailed domain-specific shaping functions~\citep{randlov1998learning}. 
Additionally, in this work, we use independently the concepts of RL approach, method, and technique to refer to the underlying learning algorithm.
These concepts have been previously used mostly equally by the RL research community.

\begin{figure}
\centering
\includegraphics[width=\linewidth]{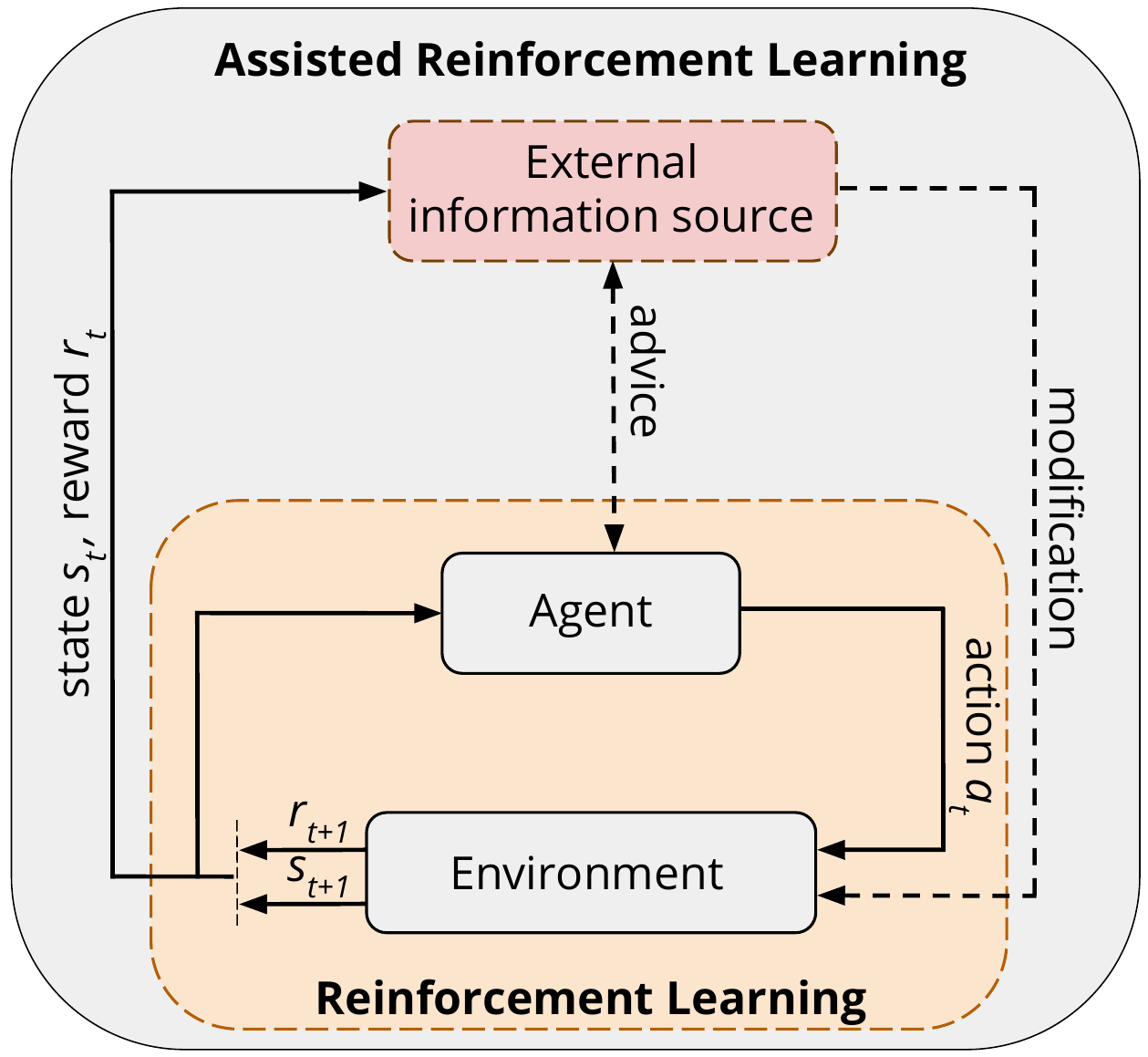}
\caption{Assisted reinforcement learning simplified framework.
In autonomous reinforcement learning, an agent performs an action $a_t$ from a state $s_t$ and the environment produces an answer leading the agent to a new state $s_{t+1}$ and receiving a reward $r_{t+1}$.
Assisted reinforcement learning adds an external information source, referred to as a trainer, teacher, advisor or assistant, that observes the environment and the agent in order to generate advice.
The trainer may advise the learner agent or sometimes directly modify the environment. 
Moreover, the agent may also actively ask advice to the external information source.
}
\label{fig:ARL_Intro}
\end{figure}

In this regard, we define \textit{Assisted reinforcement learning} (ARL) as a range of techniques that use external information, either before, during, or after training, to improve the performance of the learner agent, as well as to scale RL to larger and more complex scenarios.
While a relevant characteristic of RL is its ability to endow agents with new skills from the ground up, ARL also makes use of existing information and/or previously learned behaviour. 
Some methods for improving the agent's performance using external information include: 
directly altering weights for actions and states (biasing)~\citep{vlassis2012bayesian}; 
altering the state or action space~\citep{erez2008what}; 
critiquing past or advising on future decision-making~\citep{thomaz2007asymmetric}; 
dynamically altering reward functions~\citep{knox2010combining}; 
directly modifying the policy~\citep{griffith2013policy}; 
guiding exploration and action selection~\citep{fernandez2006probabilistic}; and, 
creating information repositories/models to supplement the environmental information~\citep{price2003accelerating}.
Figure~\ref{fig:ARL_Intro} captures all of these methods in a basic view of the ARL conceptual framework used in this work.
The classic RL approach is shown within the figure where an agent performs an action on the environment reaching a new state and obtaining a reward.
In ARL, the response of the environment is also shared with the external information source from where advice is given to the agent or changes sometimes made directly to the environment~\citep{xu2020teaching}.

To date, many methods using external information have been proposed aiming to speed up the learning process for an autonomous agent~\citep{arzate2020survey, lin2020review, dasilva2020agents, zhuang2020comprehensive}. 
Usually, they have been organized according to the technique employed, e.g., heuristic, interactive, or transfer learning, among others.
Nevertheless, there is an important lack of understanding of how these techniques are related and what characteristics they share.
Therefore, in this review, we present a conceptual framework and a taxonomy to be used to describe the practice of using external information. 
A standardised ARL taxonomy will foster collaboration between different RL communities, improve comparability, allow a precise description of new approaches, and assist in identifying and addressing key questions for further research. 


\section{A Conceptual Framework\\for Assisted Reinforcement\\Learning} \label{sec:arl_framework}

In this section, we give more details about the ARL approach including some introductory examples of works in which external information sources have been used.
Moreover, we define a conceptual framework identifying the different parts that comprise the underlying process used in ARL techniques.
Based on this conceptual framework, in the following section, we define a more detailed taxonomy for ARL approaches.

\subsection{Assisted Reinforcement Learning}

The main strength of RL is its ability for endowing an agent with new skills given no initial knowledge about the environment. 
With an appropriate reward function and enough interaction with its environment, an RL agent can learn (near-) optimal behaviour~\citep{sutton2018reinforcement}. 
The agent's behaviour at every step is defined by its policy. 
The reward function promotes desirable behaviour and sometimes penalises undesirable behaviour. 
In the traditional view of RL, the reward function, and the rewards it produces, are internal to the environment~\citep{kaelbling1996reinforcement}. 
Traditional RL, in which the environment is the sole provider of information to the agent, has been demonstrated to perform well in many different domains, especially when facing small and bounded problems~\citep{sutton2018reinforcement}. 
However, RL has some difficulties when scaling up to large, unbounded environments, particularly regarding the time needed for the agent to learn the optimal policy~\citep{cruz2016training, cruz2018multi}.
In RL, one approach to tackling this issue is to use external information to supplement the information that the environment provides~\citep{suay2011effect, millan2019human}.

Information is considered external if it originates from outside of the agent's interactions with the environment. 
In this regard, internal information is determined solely through interactions and observations with the environment. 
For example, in the case of a human the internal information would be anything the person can observe from the environment using their senses~\citep{niv2009reinforcement}.
The external information would be any information provided by peers, advisors, the internet, books, maps, and tutelage. 
In RL, anything external to the agent is usually considered part of the environment.
In this regard, if an agent is learning in an environment, a person can be considered as part of it, therefore, the agent could model that person or communicate with them~\citep{sert2020segregation}.
Although it is possible that external sources of information could be just treated as part of the environment, this is handicapping the agent in an unnecessary way.
There are external sources of information that might not necessarily be treated as part of the environment because they are socially advantaged. 
For instance, if an external source is providing action advice using directions as `left' and `right', the agent does not have to learn the meaning of these words from the ground up, or learn how to react to these instructions.
Instead, we assume the agent knows that advice is coming, what it means, and how to use it. 
For example, if a person eats some berries and later becomes sick, the person may determine that those berries are poisonous.
In this case, this would be internal information obtained by interaction with the environment. 
If instead, a peer had previously advised the person that eating those berries will make them sick, that would be external information provided by an extrinsic source. 

In this work, we refer to methods using externally-influenced agent learning as as assisted reinforcement learning. 
The ARL framework is defined to include any type of RL that uses external information to supplement agent learning and the decision-making process.
Some common practices include the direct alteration of the agent's understanding of the environment~\citep{price2003accelerating}, focusing exploration efforts through critique and advice~\citep{thomaz2007asymmetric}, or assisting the agent in the decision-making process~\citep{fernandez2006probabilistic}. 
For instance, existing ARL techniques include interactive reinforcement learning~\citep{amershi2014power, cruz2017agent}, learning from demonstration~\citep{argall2007learning, nair2018overcoming}, and transfer learning~\citep{taylor2009transfer, shao2018starcraft}, among others. 

The previously mentioned RL approaches are just examples of ARL methods that use external information to supplement the agent's decision-making process and learning. 
Additional details of these and other approaches and how they use an external information source to assist the agent (in terms of our ARL framework) are addressed in Section~\ref{sec:arl_examples}. 
The external information source is most commonly a human or another artificial agent. 
Regardless of the source, the use of external information has often been shown to improve an agent's ability and learning speed. 
In the next section, we present a more detailed conceptual framework for ARL which is the base for the taxonomy we propose subsequently.

\subsection{Conceptual Framework}
\label{subsec:arl_framework}

\begin{figure*}[t!]
\centering
\includegraphics[width=0.7\linewidth]{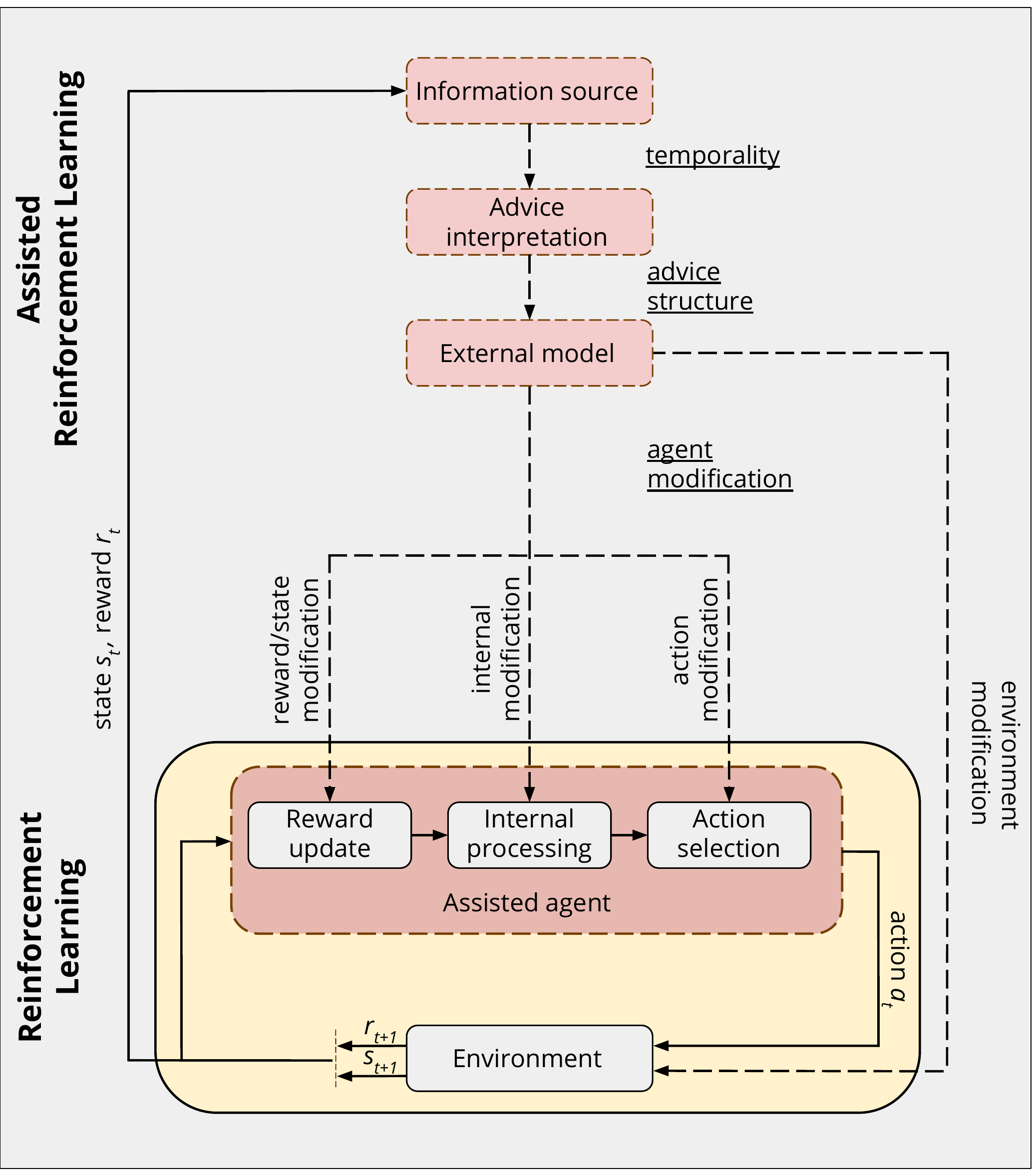}
\caption{Detailed view of the assisted reinforcement learning framework.
The diagram includes four processing components shown as dashed red boxes.
Inside the assisted agent, one can observe three different points where it can receive possible modifications from the external model.
Additionally, three communication links are shown with underlined text.
This framework is subsequently used to further discuss the proposed ARL taxonomy.
}
\label{fig:ARL_Full}
\end{figure*}

The proposed ARL framework is built to improve the classification, the comparability, and the discussion on different externally-influenced RL methods. 
To achieve this aim, the framework has been designed using insights and observations drawn from many different ARL approaches. 
The result is a framework that can describe existing methods while also being flexible enough to include future research. 
The framework details are shown in Figure~\ref{fig:ARL_Full}. 

The proposed ARL framework comprises four processing components shown using red boxes in the diagram, i.e., information source, advice interpretation, external model, and the assisted agent itself.
The external information source may not have perfect observability and also may not know details about the RL agent (algorithms, weights, hyperparameters, etc.), or make assumptions, e.g., value-based learners~\citep{taylor2005value}. 
The processing components are responsible for providing, transforming, and storing information. 
We do include the agent as part of the processing components since it is part of the RL process as well. 
However, an agent using ARL generally behaves as a traditional RL agent, i.e., it interacts with the environment by exploring/exploiting actions.
Inside the agent, there are three different stages: reward update, internal processing, and action selection.
Each of those stages may be altered by the external model using reward/state modifications, internal modifications, or action modifications respectively.
Moreover, the ARL framework also comprises three communication links that connect the four processing components and are labelled: temporality, advice structure, and agent modification.
These links are shown between the processing components and represent the communication lines in Figure~\ref{fig:ARL_Full} that connect the processing components together.
The communication links convey information or denote constraints on the data such as where or when to provide information.

The ARL framework describes the transmission, modification, and modality of sourced information. 
In this regard, we consider the ARL framework as a whole unit, comprising traditional autonomous RL plus the components and links for assistance.
Thus, the taxonomy is a part of the framework and oriented to describe the assisted learning section.
Although the framework has been developed on how ARL is usually built, not all ARL approaches use all the proposed components and links.
Below, we briefly describe each of the components and links of the framework.
They are subsequently used in the next section to describe in detail the proposed taxonomy.

\begin{itemize}
\item \textbf{Information source}: is the origin of the assistance being provided to the agent.
The source may be a human, a repository, or another agent.
There may be multiple information sources providing assistance to an agent.
\item \textbf{Temporality}: determines both the time at which information is provided to the agent, and the frequency with which it is provided.
Information may be provided, before, during, or after agent training, and occur multiple times through the learning process. 
Therefore, it is also responsible for how the information source communicates temporal issues to the advice interpreter. 
\item \textbf{Advice interpretation}: denotes the process of transforming incoming information into a format better suited for the agent.
This may involve extracting key frames from video, converting audio samples to rewards, or mapping information to states.
\item \textbf{Advice structure}: represents the structure of the advice after translation in a form suitable for the external model. 
Some approaches may not have an explicit external model, therefore, this structure might instead be directly used to modify the agent. 
\item \textbf{External model}: is responsible for retaining and relaying the information between the source and the agent. 
The model may retain the received information in the learning model, using it for later decisions, or it may discard the received information as soon as it has been used. 
\item \textbf{Agent modification}: denotes the approach that the agent uses to benefit from the incoming information. 
The most common modification approaches may use information to alter the environmental reward signal or modify the agent's behaviour or the decision-making process directly.
\item \textbf{Assisted Agent}: is the RL agent receiving the external information or advice while learning a new task. 
The agent needs to work out how to incorporate the provided information with its own learning. 
If a different action is suggested by the trainer then the agent may decide if it should follow to that advice or not.
\end{itemize}

\begin{figure}
\centering
\includegraphics[width=\linewidth]{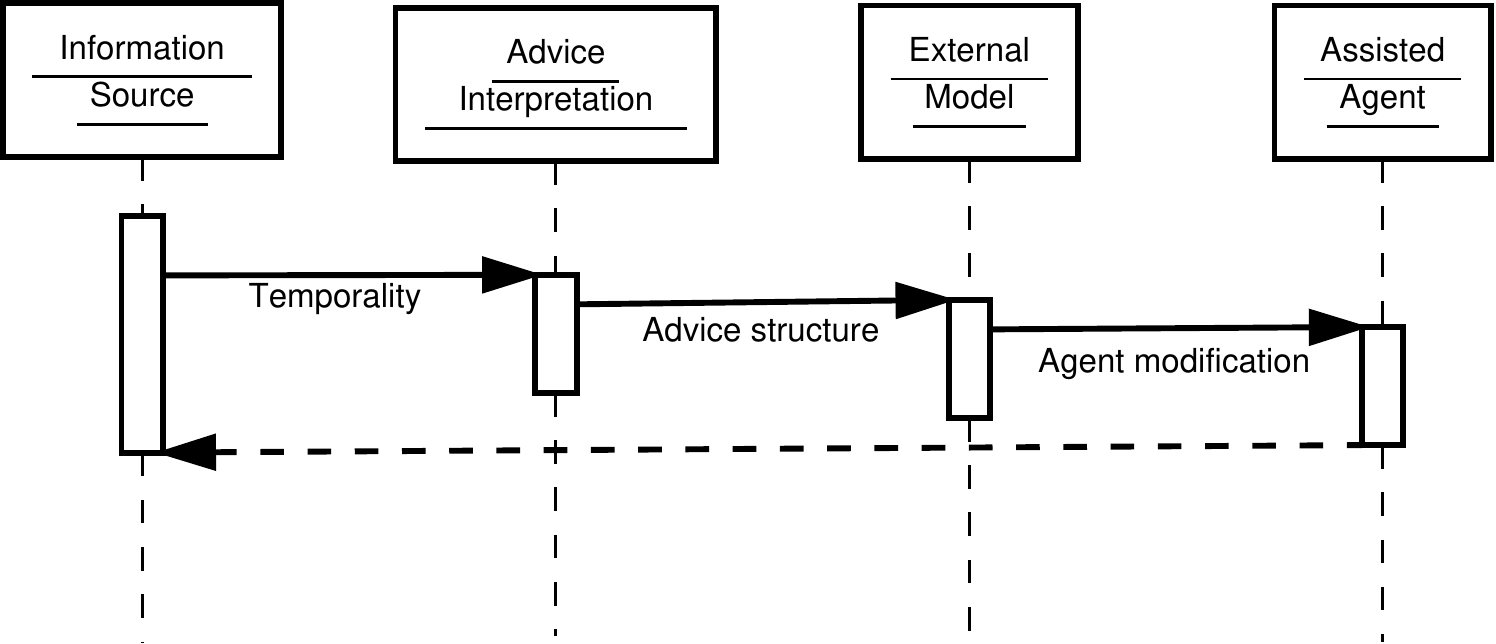}
\caption{
Relation between the processing components and the communication links as a UML sequence diagram.
}
\label{fig:UML_Sequence}
\end{figure}

Figure~\ref{fig:UML_Sequence} shows in a UML sequence diagram the interaction between the processing components and communication links according to Figure~\ref{fig:ARL_Full}.

\section{Assisted Reinforcement\\Learning Taxonomy} \label{sec:arl_taxonomy}

\begin{figure}[t!]
\centering
\includegraphics[width=\linewidth]{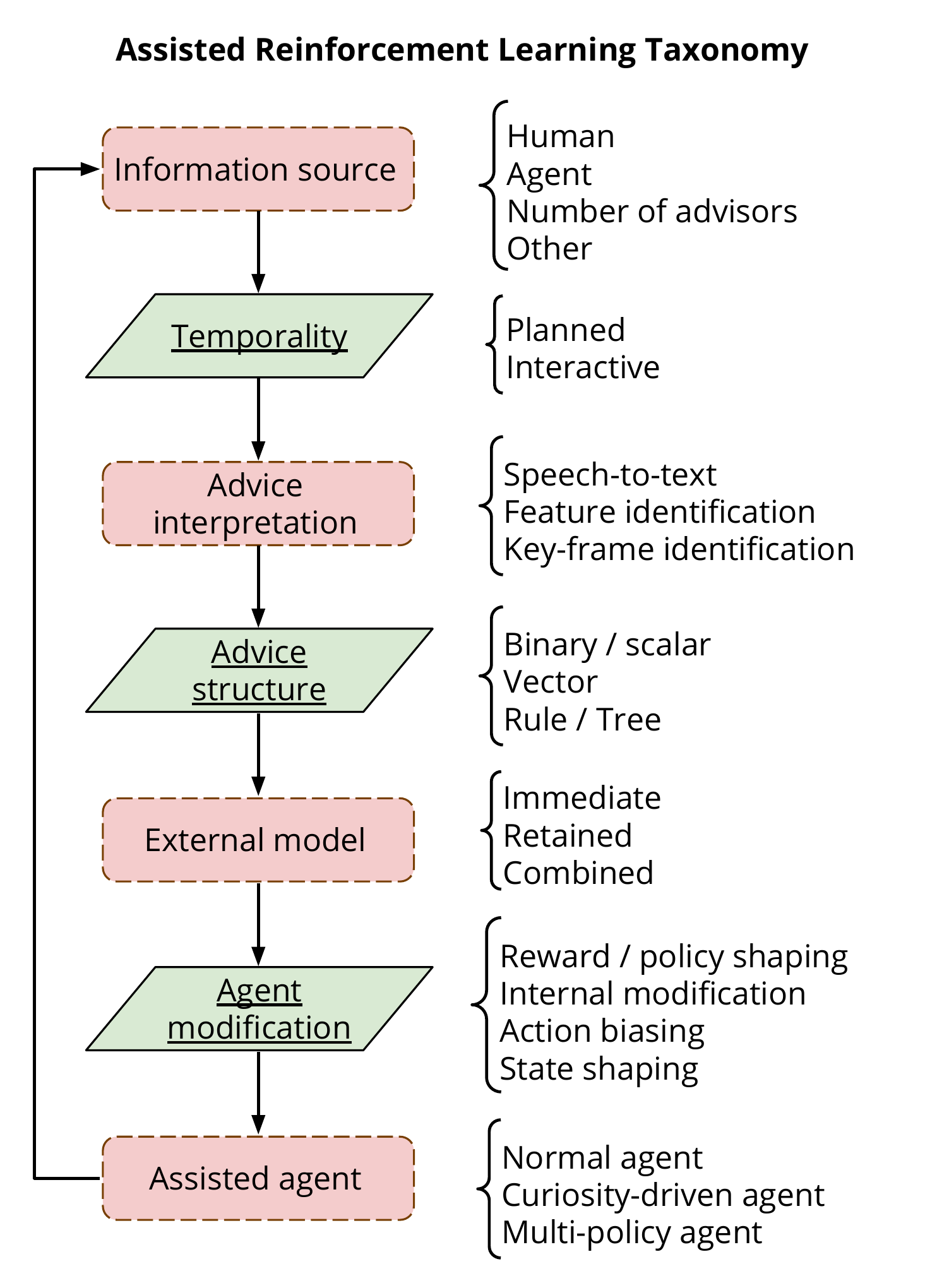}
\caption{The assisted reinforcement learning taxonomy.
This figure shows the four processing components as dashed red boxes and the communication links as green parallelograms using underlined text.
Examples for each component and method are included at the right.
}
\label{fig:ARL_Taxo}
\end{figure}

In this section, we describe the processing components and communication links included in the proposed framework within an ARL taxonomy\footnote{In this context, we refer the taxonomy as a classification of the different elements of the ARL framework, i.e., processing components and communication links, and not as a way to classify each ARL method.} and give more details of each of them.
Figure~\ref{fig:ARL_Taxo} shows all the elements of the proposed ARL taxonomy including examples for each processing component and communication link.
In the taxonomy, we include the agent as a component being the one that receives the advice.
Each of the seven elements, i.e., processing components and communication links, is described in detail in the following subsections.
In our work, the concept of taxonomy is used to classify the different elements within a class of problems, i.e., ARL problems. 
In this regard, our proposal is represented by a general ontology where the class is ARL, the properties are the processing components and the communication links, and the relations between the properties are as shown in Figure~\ref{fig:ARL_Taxo}. 

\subsection{Information Source}
\label{subsec:information_source}
The external information source is the main factor that sets ARL apart from traditional RL approaches. 
It is responsible for introducing new information about the task to the agent, supplementing or replacing the information the agent receives from the environment. 
The source is external to the agent and the environment, providing information that either the agent may not have had access to, or would have eventually learned itself. 
The information source may be able to observe the environment, the agent, or the agent's decision-making process. 
The objective of the information source is to assist the agent in achieving its goal faster. 

There may be multiple information sources communicating with an agent. 
This may be humans, agents, other digital sources, or any combination of the three~\citep{isbell2000cobot}. 
The use of multiple sources offers a wider range of available information to the agent. 
However, more complex modification methods may be required to manage the information and handle conflicting advice~\citep{kamar2012combining}.

There are many examples of external information sources in current ARL literature, the most common of which are humans and additional reward functions~\citep{ng1999policy, thomaz2006reinforcementAAAI, millan2019human}. 
For instance, RLfD and IntRL use human guidance to provide the agent with a generalised view of the solution~\citep{cobo2014abstraction, subramanian2016exploration}. 
Moreover, the use of additional reward functions is one of the earliest examples of ARL. 
In such cases, the designer of the agent encodes some further information about the environment or goal as an additional reward, supplementing the original reward given by the environment. 

An example of the use of additional reward functions can be found in Randl\o v and Alstr\o m's bicycle experiment~\citep{randlov1998learning}, in which, they teach an agent to ride a bicycle towards a goal point. 
Without additional assistance, the RL agent would only receive a reward upon reaching the termination state. 
Randl\o v and Alstr\o m encoded some of their knowledge as a shaping reward signal external to the environment, providing the agent with additional rewards if it is cycling towards the goal point. 
In this scenario, the system designers acted as an external information source, providing extra information to the RL agent. 
The use of this external information results in the agent learning the solution faster than using the traditional RL approach.

Some other information sources include behaviours from past experiences or other agents, repositories of labelled data or examples, or distribution tables for initialising/biasing agent behaviour~\citep{cruz2017agent}.
Video, audio, and text sources may be used as well~\citep{cruz2016multi}. 
However, these sources may require substantial amounts of interpretation and preprocessing to be of use.

The accuracy, availability, or consistency of the information source can affect the maximum utility of the information~\citep{torrey2013teaching, taylor2014reinforcement}. 
Identifying in advance inaccurate information given to the agent can significantly improve performance~\citep{cruz2016training, cruz2018improving}.
While the information source may perform the validation and the verification of the given advice, the primary duty remains simply to act as a supplementary source of information. 
In this regard, both validation and verification of information are functions better suited for the external model or the assisted agent.

\subsection{Temporality}
\label{subsec:temporal}
The temporal component, or temporality, refers to the time at which information is communicated by the information source. 
The information may be provided in full to the agent at a set time (either before, during, or after training). 
This is referred to as \textit{planned assistance}~\citep{partalas2008reinforcement, cheng2013framework}. 
Alternatively, the information may be provided at any time during the agent's operation, referred to as \textit{interactive assistance}~\citep{pilarski2012between, stahlhut2015interaction}.

Planned assistance, on the one hand, is common in ARL methods. 
Some examples are predefined additional shaping functions, agent policy initialisation based on either prior experience or a known distribution, and the creation of subgoals that lead the way to a final solution~\citep{partalas2008reinforcement}. 
These methods let the experiment designer endow the agent with initial information about the environment or the goal to be achieved. 
By providing this initial knowledge, the designer can reduce the agent's need for exploration. 

The bicycle experiment discussed in the previous section is an example of planned assistance. 
As mentioned, the agent is learning to control a bicycle and must learn to steer it towards a goal~\citep{randlov1998learning}. 
Before the experiment, the designers give the agent additional information in the form of a reward signal that correlates to the direction of the goal state. 
This planned assistance approach helps the agent to narrow the search space by giving it extra information about the environment. 
This small yet beneficial initial information results in a significant improvement in the agent's learning speed.

Another example of planned assistance is found in heuristic RL. 
Heuristic RL is a method of applying advice to agent decision-making. 
One example is an experiment which implements heuristic RL in the RoboCup soccer domain~\citep{celiberto2007heuristic}, a domain known for its large state space and continuous state range. 
In this environment, one team attempts to score a goal, while the other team tries to block the first team from scoring, such as in half-field offence~\citep{kalyanakrishnan2006half, hausknecht2016half}. 
In this experiment using heuristic RL, the defending team is given initial advice before training. 
This advice consists of two rules: if the agent is not near the ball then move closer, and if the agent is near the ball then do something with it. 
The experiment results show that a team that uses planned assistance performs better than a team that is given no initial knowledge~\citep{celiberto2007heuristic}.

Interactive assistance, on the other hand, refers to information provided by the source repeatedly throughout the agent's learning. 
Information sources that assist interactively often can observe the agent's current state, or the environment the agent is operating in. 
In current literature, humans are more commonly used as information sources for interactive assistance~\citep{thomaz2006reinforcementROMAN, subramanian2011learning}. 
The human can observe how the agent is performing and its current state in the environment, and provides guidance or critiques of the agent's behaviour~\citep{bignold2020human}.

For example, Sophie's Kitchen~\citep{thomaz2007asymmetric} presents an IntRL based agent, called Sophie, which attempts to bake a cake by interacting with the items and ingredients found in a kitchen. 
In this experiment, the agent will receive a reward if it successfully bakes the cake.
At any point during the agent's training, an observing human can provide the agent with an additional reward to supplement the reward signal given by the environment. 
If the agent performs an undesirable action, such as forgetting to add eggs to the cake, the human can punish the agent by providing an immediate negative reward. 
The human can also reward the agent for performing desirable actions, such as adding ingredients in the correct order. 
In this experiment, the human advisor is acting as an interactive information source. 

Although the agent could learn the task without any assistance, the addition of the human advisor and interactive feedback allows the agent to learn the desired behaviour faster in comparison to autonomous RL~\citep{thomaz2007asymmetric}. 
The benefit of using interactive advice rather than planned advice is that the information source can react to the current state of the agent. 
Additionally, an interactive information source does not need to encode all possibly useful advice up front. 
Instead, it can choose to provide relevant information only when required. 
This approach does have a significant cost; the information source needs to be constantly observing the agent and determining what information is relevant. 
For instance, an approach using inverse RL through demonstrations may also consider providing failed examples to show the agent what not to do~\citep{shiarlis2016inverse}.

\subsection{Advice Interpretation}
\label{subsec:interpretation}
The advice interpretation stage of the taxonomy denotes what transformations need to occur on the incoming information. 
The source provides information for the agent to use that may need to be translated into a format that the agent can understand. 
The information source may provide their assistance in many different forms. 
Some examples include audio~\citep{cruz2015interactive}, video~\citep{cruz2016multi}, text~\citep{liu2019privacy}, distributions and probabilities~\citep{millan2019human}, or prior learned behaviour from a different task or agent~\citep{dasilva2020agents}. 
This information needs to be adapted for use by the agent for the current task. 
The product of the advice interpretation stage depends on the structure that the agent or external model requires.

A field where the interpretation of incoming advice is crucial is Transfer Learning (TL). 
The goal of TL is to use behaviour learned in a prior task to improve performance in a new, previously unseen task~\citep{dasilva2019survey}. 
A critical step in TL is the mapping of states and observations between the old and new domains. 
The information source provides information to the agent that does not fully align with its current task. 
Therefore, it is crucial that the information provided can be correctly interpreted, so as to be useful to the current domain. 
More commonly, this interpretation stage in TL is performed by hand. However, there has also been effort attempting to automate this stage~\citep{taylor2008autonomous, narvekar2016source}.

Another example of the use of the advice interpretation stage is with the sourcing of feedback for RL agents. 
In the Sophie's Kitchen experiment~\citep{thomaz2007asymmetric}, discussed in the previous section, the agent can be given positive or negative feedback by a human regarding its choice of actions. 
In this experiment, the human creates either a green (positive) or a red (negative) bar to represent the desired feedback to be given to the agent. 
This bar is used to interpret the reward signal to give to the agent, with the colour of the bar designating whether the reward is positive or negative, and the size of the bar designating the magnitude of the reward. 
This type of feedback can also be extended to audio, where recording phrases such as `Good' or `Well Done' are interpreted as positive rewards and `Bad' or `Try Again' are interpreted as negative rewards~\citep{tenorio-gonzalez2010dynamic}. 

These methods can also be combined into a multi-model architecture to provide advice to an RL robotic agent using audiovisual sensory inputs, such as work by Cruz et al.~\citep{cruz2016multi}. 
In this experiment, a simulated robot learns how to clean a table using a multi-modal associative function to integrate auditory and visual cues into a single piece of advice which is used by the RL algorithm.
In this scenario, the external information source is a human trainer and the RL algorithm represents the integrated advice as a state-action pair.

\subsection{Advice Structure}
\label{subsec:structure}
The advice structure component refers to the form that the agent or external model requires incoming information to take. 
The information that the agent uses can be represented in a number of ways. 
Some examples of advice structures include: Boolean values denoting positive or negative feedback; rules determining action selection; matrices for mapping prior experiences to new states; case-based reasoning structures for the agent to consult with; or, hierarchical decision trees to represent options for the agent to take~\citep{subramanian2011learning, kaplan2002robotic}. 

The simplest form of structure is binary, in which the information takes only one from two options, such as `Good' or `Bad'. 
An example of the use of a binary structure is the TAMER-RL agent~\citep{knox2009interactively}. 
TAMER-RL is an IntRL agent that uses binary feedback from an observing human. 
At any time step, the human can agree or disagree with the agent about its last action. 
In this case, the feedback is a binary structure indicating agree or disagree. 

A more complex advice structure is used in case-based RL agents~\citep{sharma2007transfer}. 
A case in this context represents a generalised area of the state space and provides information about which actions to take in that state. 
The use of a case-based structure allows the agent to gain more information from the information source compared to a binary structure, at a cost of more complex sourcing and interpretation approaches.

One of the more common advice structures is a simple state-action pair. 
A state-action pair consists of a single state and an associated piece of advice. 
The associated advice may be an additional scalar reward or a recommended action. 
Using a state-action pair, sourced information is interpreted to provide advice for a given state. 
In the cleaning-table robot task~\citep{cruz2016multi}, discussed in the previous section, the external trainer using multi-modal advice provides an action to be performed in specific states.
Once the advice is processed using the multi-modal integration function, the proposed action is given to the RL agent to be executed as a state-action pair considering the agent's current state. 
This state-action structure has also been used for other methods including TAMER-RL~\citep{knox2009interactively}, Sophie's Kitchen~\citep{thomaz2007asymmetric}, and policy-shaping approaches~\citep{griffith2013policy}. 

A novel rule-based interactive advice structure is introduced in~\citep{bignold2021persistent}.
Interactive RL methods rely on constant human supervision and evaluation, requiring a  substantial commitment from the advice-giver. 
This constraint restricts the user to providing advice relevant to the current state and no other, even when such advice may be applicable to multiple states.
Allowing users to provide information in the form of rules, rather than per-state action recommendations, increases the information per interaction, and does not limit the information to the current state.
Rules can be interactively created during the agent's operation and be generalised over the state space while remaining flexible enough to handle potentially inaccurate or irrelevant information.  
The learner agent uses the rules as persistent advice allowing the retention and reuse of the information in the future.
Rule-based advice significantly reduces human guidance requirements while improving agent performance.

\subsection{External Model}
The external model is responsible for retaining and relaying information between the information source and the agent. 
The external model receives interpreted information from the information source and may either retain the information for use by the agent when required or pass it to the agent immediately. 

A \textit{retained model} is an external model that stores all information provided by the information source~\citep{fernandez2006probabilistic}. 
A retained model may be used if the cost of acquiring information is greater than the cost of storing it, if the information provided is general or applies to multiple states, or if the information is gathered incrementally. 
In instances where information is gathered incrementally, using a retained model allows the agent to build up a knowledge base over time. 
The agent may consult with the model at any time to determine if a reward signal is to be altered, or if there is any extra information that may assist with decision-making.

An \textit{immediate model} passes the information directly to the agent~\citep{moreira2020deep}.
In this case, the information received is only relevant to the current time step, or the cost of reacquiring the information from the source is less than that of retaining the information. 

Approaches can also combine this by incorporating both a retained model as well as passing some information through directly, such as~\citep{cruz2016training}.
In this work, an RL agent uses a combination of interactive feedback and contextual affordances~\citep{cruz2016learning} to speed up the learning process of a robot performing a domestic task.
On the one hand, contextual affordances are learned at the beginning of autonomous RL and are readily available from there on to avoid the so-called failed-states, which are states from where the robot is not able to finish the task successfully anymore.
On the other hand, interactive feedback is provided by an external advisor and used to suggest actions to perform when the robot is learning the task.
This advice is given to the robot to be used in the current state and it is discarded immediately after.


The external model may have different functions depending on its implementation. 
For instance, heuristic RL hosts a model that stores rules and advice that generalise over sections of the state space~\citep{dorigo2014antq}. 
In TL, the external model may hold information regarding past experiences and policies from problems similar to the current domain~\citep{taylor2009transfer, banerjee2007general}, or in inverse RL, the external model is a substitute for the reward function~\citep{abbeel2004apprenticeship}. 

\subsection{Agent Modification}
The modification stage of the framework denotes how the information that the external model contains is used to assist the agent in achieving its goal. 
It is responsible for supplementing the agent's reward, altering the agent's policy, or helping with the decision-making process. 
A popular method for injecting external information into agent learning is shaping~\citep{skinner1975shaping}.  
Shaping is a common method for altering agent performance by modifying parameters in the learning process. 
Erez and Smart~\citep{erez2008what} propose a list of techniques in which shaping can be applied to RL agents. 
These include altering the reward, the agent's policy, agent learning parameters, and environmental dynamics~\citep{xu2020teaching}.

Altering the reward the agent receives is a straightforward method for influencing an agent's learning~\citep{churamani2020icub}. 
It is known as reward-shaping, in which the external information is used to bias the agent's learning~\citep{ng1999policy}. 
Special care must be taken to ensure that any modification of the reward signal remains zero-sum to avoid the agent exploiting the shaped reward in ways that do not align with the desired goal. 
This can be achieved by ensuring that additional rewards are potential-based, meaning that they are derived from the difference in the values of a potential function at the current and successor states~\citep{harutyunyan2015expressing}. 
However, recent work by~\citep{paniz2020useful} shows a flaw in the previous method when transforming non-potential-based reward-shaping into potential-based. 
Alternatively, the authors introduce a policy invariant explicit shaping algorithm allowing for arbitrary advice, confirming that it ensures convergence to the optimal policy when the advice is misleading and also accelerates learning when the advice is useful~\citep{paniz2020useful}.
Shaping techniques have also been used to alter state-action pairs~\citep{wiewiora2003principled}, for dynamic situations~\citep{harutyunyan2015expressing, devlin2012dynamic}, and for multi-agent systems~\citep{devlin2011theoretical}.

Policy-shaping is the modification of the agent's behaviour~\citep{griffith2013policy}. 
This modification can be done either by influencing how the agent makes decisions or by directly altering the agent's learned behaviour. 
A simple method of policy-shaping involves forcing it to take certain actions if advice from the information source has recommended them~\citep{grizou2013robot, navidi2020human}. 
Human-in-the-loop techniques may be beneficial to address complex RL problems with the help of domain experts, e.g., in health informatics~\citep{holzinger2016interactive}.
This allows the external information source to guide the agent and take direct control over exploration/exploitation. 
Alternatively, the information source can choose to alter the agent's behaviour directly by changing Q-values or installing rules that override the actions for chosen states~\citep{knowles2008hybrid}.
This method of modification can improve agent performance rapidly, as it can give the agent partial solutions. 

Internal modification is a method of altering the parameters of the agent that are essential to its learning. 
Parameters such as the learning rate ($\alpha$), discount factor ($\gamma$), and exploration percentage ($\epsilon$), are all internal to the RL agent and may be altered to affect its performance~\citep{tesauro2004extending}.
For example, if an advisor observes that an agent is repeating actions and not exploring enough then the exploration percentage or learning rate may be temporarily increased. 
Internal modification is a simple method to implement. 
However, it can be difficult at times to know which parameters to adjust, and to what degree they are to be adjusted.

Environmental modification is an indirect method for influencing an RL agent. 
Altering the environment is not always achievable and may be a technique better suited for digital or simulated environments. 
Some examples of modifying the environment include altering or reducing the state space and observable information~\citep{kerzel2018accelerating, breyer2019comparing}, reducing the action space~\citep{sridharan2017can}, modifying the agent's starting state~\citep{dixon2000incorporating}, or altering the dynamics of the environment to make the task easier to solve~\citep{millan2021robust} 
Below, we further describe these environmental modifications.

Reducing the state space can speed up the agent's learning as there is less of the environment to search. 
While the agent cannot fully solve the task with an incomplete environment representation, it allows the agent to learn the basic behaviour. 
The level of detail in the state representation can then be increased, allowing the agent to refine its policy towards the correct behaviour~\citep{kerzel2018accelerating, breyer2019comparing}.
Reducing the action space is similar to the previous. 
The agent's available actions are limited, and the agent attempts to learn the best behaviour it can with the actions it has available. 
Once a suitable behaviour has been achieved, new actions can be provided, and the agent can begin to learn more complex solutions~\citep{sridharan2017can}.
Modifying the agent's starting space alters where in the environment the agent begins learning. 
Using this approach, the agent can begin training close to the goal. 
As the agent learns how to navigate to the goal, the starting state is incrementally moved further away. 
This allows the agent to build upon its past knowledge of the environment~\citep{dixon2000incorporating}. 
Altering the dynamics of the environment involves changing how the environment operates to make the task easier for the agent to learn~\citep{xu2020teaching}. 
By altering attributes of the environment such as reducing gravity, lowering maximum driving speed, or reducing noise, the agent may learn the desired behaviour faster or more safely. 
After the agent learns a satisfactory behaviour, the environment dynamics can be changed to more typical levels~\citep{millan2020robust}.

\subsection{Assisted Agent}
\label{subsec:agent}
The final component of the proposed ARL taxonomy is the RL agent. 
A key aspect of the taxonomy is that the agent, in the absence of any external information, should operate the same as any RL agent would. 
Given no external information, the agent should continue to explore and interact autonomously with its environment and attempt to achieve its goal.

In the next section, we present an in-depth look at some ARL techniques and describe them in terms of the taxonomy that has been presented in this section.

\section{Illustrative Approaches with\\Components and Links from\\the Taxonomy}\label{sec:arl_examples}

This section presents an in-depth analysis of some popular and well-known ARL approaches. 
Each illustrative approach is described as an instance of the proposed taxonomy shown in Section~\ref{sec:arl_taxonomy}, in some cases using a specific approach and in other cases a set of them.
Therefore, for each presented ARL approach, we show how each processing component and each communication link particularly adapts to the ARL taxonomy using current literature in the respective field for concrete examples.

\subsection{Heuristic Reinforcement Learning}

Heuristic RL uses pieces of information that generalise over an area of the state space. 
The information is used to assist the agent in decision-making and reduce the searchable state space~\citep{bianchi2015transferring, yang2019emergency}.
An example of a heuristic is a rule. 
A rule can cover multiple states, making its use efficient at delivering advice to an agent. 
In Section~\ref{subsec:temporal}, we have introduced a heuristic RL experiment applied to the RoboCup soccer domain~\citep{celiberto2007heuristic}.
In the RoboCup soccer domain, one team actively tries to score a goal, while the other team tries to block it. 
As mentioned, the defending team is given initial advice before training, consisting of two predefined rules. 
The following is an analysis of this heuristic RL example applied as the ARL taxonomy.

\begin{figure}[t!]
\centering
\includegraphics[width=\linewidth]{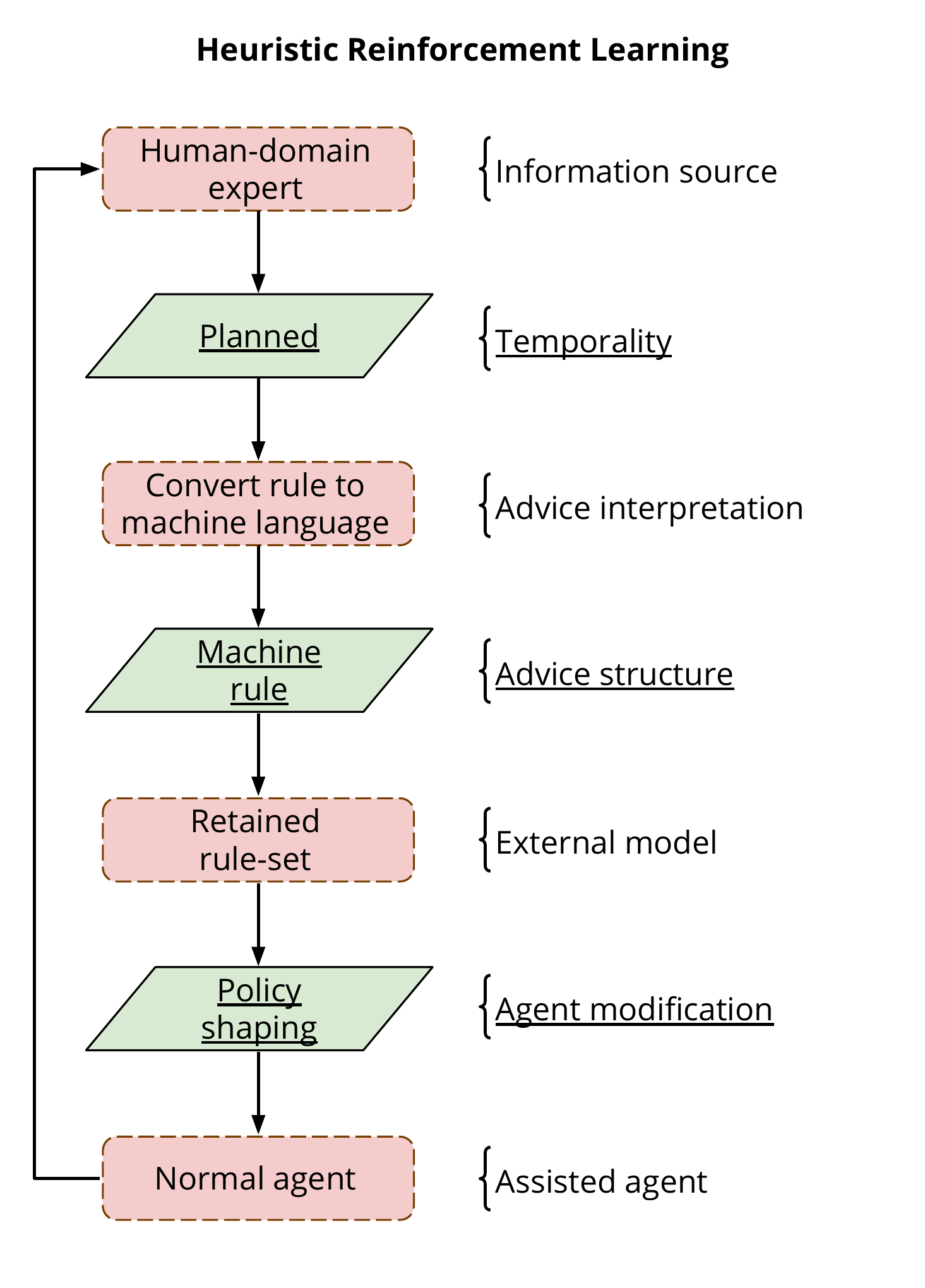}
\caption{Heuristic RL components according the proposed ARL taxonomy. 
The particular processing components and communication links illustrate a technique used in the RoboCup soccer domain~\citep{celiberto2007heuristic}.}
\label{fig:ARL Heuristic Reinforcement Learning}
\end{figure}

\begin{itemize}
\item \textbf{Information source}: 
The information source for the RoboCup experiment is a person. 
In this case, the person has previously experimented with the robot soccer domain and can advise the agent with some rules that will speed up learning.

\item \textbf{Temporality}: 
The advice for the agent is given before training begins. 
Once training has begun the person does not interact with the agent again. 
This is an example of planned assistance, where information is given to the agent at a fixed time, and the information is known by the information source in advance.

\item \textbf{Advice interpretation}: 
The information needs to be understandable by the agent. 
In the robot soccer domain, the person gives two rules; (i) if not near the ball then move towards the ball, and (ii) if near the ball do something with the ball. 
These rules are understandable by the human but need to be translated into machine code so that agent can use them. 
This is usually a task easily performed by a knowledgeable human operator. 
The result is conditional-like rules as: (i) \textbf{IF NOT} close\_to\_ball() \textbf{THEN} target\_and\_move(), and (ii) \textbf{IF} close\_to\_ball() \textbf{THEN} kick\_ball().

\item \textbf{Advice structure}:
The structure of the advice after being interpreted is a new rule. 
The rule needs to be compatible with the agent, including the ability to substitute variables and evaluate expressions. 

\item \textbf{External model}: 
The external model used by the heuristic RL agent is a rule set. 
The external model retains all rules given to it. 
The model may also retain statistics about the rule relating to confidence, number of uses, and state space covered.

\item \textbf{Agent modification}: 
Heuristic RL uses the rule set to assist the agent in its decision-making. 
If a rule applies to the current state, then the action that the rule recommends is taken by the agent. 
This is a form of policy-shaping as the agent's decision-making is directly manipulated by the external information.

\item \textbf{Assisted Agent}: 
The RL agent operates as usual. 
When it is time to decide on an action to take it consults the external model. 
The external model tests all the rules it has and checks to see if any applies to the current state, otherwise, the agent's default decision-making mechanism is used.
\end{itemize}

Figure~\ref{fig:ARL Heuristic Reinforcement Learning} shows how the heuristic RL approach fits into the proposed ARL taxonomy taking into consideration the previous definitions of processing components and communication links from the RoboCup soccer domain.

\subsection{Interactive Reinforcement Learning}

IntRL is another application of ARL. 
Most commonly, the information source is an observing human or a substitute for a human, such as an oracle, a simulated user, or another agent~\citep{thomaz2005real}. 
The human provides assessment and advice to the agent, reinforcing the agent’s past actions and guiding future decisions. 
The human can assess past actions in two ways, by stating that the agent's chosen action is somehow correct or incorrect, or by telling the agent what the correct action to take is in that instance. 
Alternatively, the human can advise the agent on what actions to take in the future~\citep{li2019human}.
The human can recommend actions to take or to avoid, or provide more information about the current state to assist the agent in its decision-making~\citep{cruz2018multi}.

IntRL applications include having a human to provide additional reward information~\citep{knox2012reinforcementAAMAS, knox2012reinforcementROMAN}, and having a human or agent provide action advice~\citep{zhan206theoretically,amir2016interactive}.
All of these methods work in real-time and similarly, differing mainly in the agent modification stage.  
The following is an analysis of these IntRL approaches applied as the ARL taxonomy.

\begin{itemize}
\item \textbf{Information source}: 
The information source is a human or simulated user. 
A simulated user is a program, analogous to a human, that acts how a human would in a given situation. 
The human can observe the agent's current and past states, past actions taken, and what action the agent recommends it takes~\citep{bignold2021evaluation}.

\item \textbf{Temporality}: 
IntRL agents operate interactively. 
The advisor can provide information to the agent before, during, or after learning, and repeatedly throughout the learning process. 
This allows the advisor to react to current information and supply the agent with relevant advice.

\item \textbf{Advice interpretation}: 
The advisor provides either an assessment of past actions taken, recommendations about actions to take, or a reward signal. 
Computer simulated agents can receive this information as key presses. 
However, physical agents may receive this information through audio or video inputs~\citep{cruz2016multi}. 
In the case of audio inputs, these may be simple commands such as ’Correct’ or ’Go Right’, which can be translated to a form the agent can understand~\citep{cruz2015interactive}.
Supporting input modalities such as natural language makes systems based on IntRL more accessible to users who are not themselves familiar with RL. 

\item \textbf{Advice structure}: 
A common structure of advice the agent requires is simply a state-action pair. 
Using this structure the human can assign advice to a state for the agent to use, such as: In this state, do this~\citep{ayala2019reinforcement}.

\item \textbf{External model}: 
Either retained or immediate models are commonly used~\citep{fernandez2006probabilistic, knox2012humans}.
A retained model tracks what advice/feedback has been received for each state~\citep{fernandez2006probabilistic}. 
The agent can use this model to determine the human’s accuracy, consistency, and discount for each piece of advice received. 
The model acts as a lookup table for the agent, if advice exists for the current state, then the agent can use it. 
Alternative methods may not retain information given by the human and only use it for the current state~\citep{knox2012humans}. 

\item \textbf{Agent modification}: 
The most common methods of using the advice to modify the agents learning process are reward- and policy-shaping~\citep{li2019human}. 
Reward-shaping uses assessment/critique gathered from the advisor to alter the reward given to the agent. 
If the advisor disagrees with a past action, then the reward received for that state-action pair is decreased. 
If the advisor recommends an action to take in the future, then policy-shaping can be used to override the agent’s usual action selection mechanism. 
One method of implementing policy-shaping for interactive advice is probabilistic policy reuse~\citep{fernandez2006probabilistic}.

\item \textbf{Assisted Agent}: 
Most of the time, the RL agent operates as any other RL agent would, i.e., it performs actions in the environment by exploiting/ exploring.
The agent should continue to do so even if no advice from the trainer is given.
Although a trainer could proactively provide advice to the learner, sometimes the student could decide to request such advice, and the trainer may or may not respond to that request. 
For instance, heuristics have been used to decide if the trainer should provide advice and/or if the learner should ask for it~\citep{amir2016interactive}.
In contrast, recent work estimates the learner's uncertainty in its current state, asking for advice in case the level of uncertainty is above a predefined threshold~\citep{dasilva2020uncertainty}.
\end{itemize}

\begin{figure}[t!]
\centering
\includegraphics[width=1\linewidth]{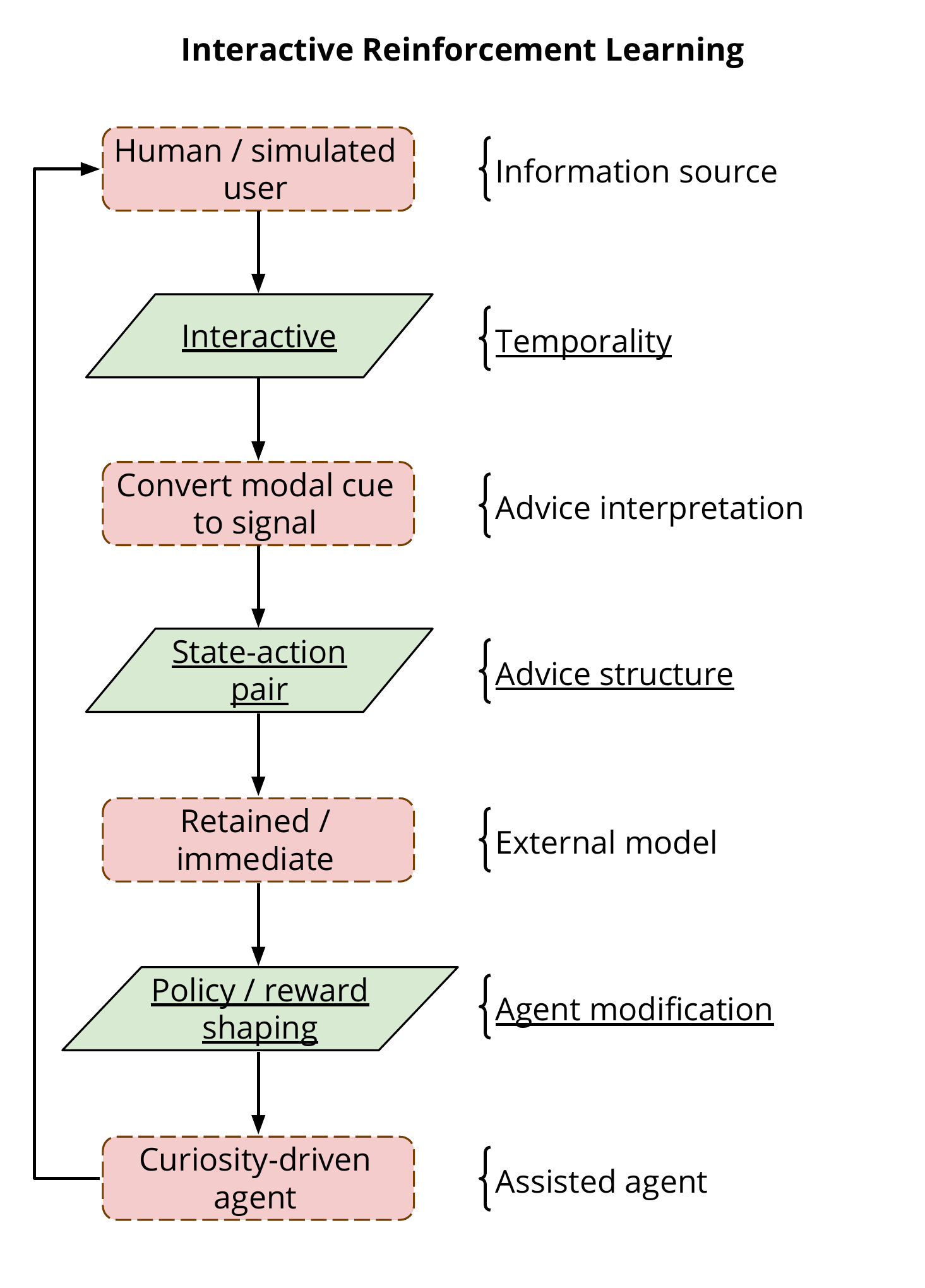}
\caption{Interactive RL as the proposed ARL taxonomy. In this approach, interactive advice is given by the user and more commonly used as policy and reward shaping.}
\label{fig:ARL_Interactive Reinforcement Learning}
\end{figure}

Figure~\ref{fig:ARL_Interactive Reinforcement Learning} shows how the IntRL approach is adapted to the proposed ARL taxonomy taking into account the previous definitions of processing components and communication links.

\subsection{Reinforcement Learning from Demonstration}
RLfD is a term coined by Schaal~\citep{schaal1997learning}. 
It refers to the setting where both a reward signal and demonstrations are available to learn from, combining the best of the fields of RL and Learning from Demonstration (LfD). 
Since RL presents an objective evaluation of behaviour, optimal behaviour can be achieved. 
Such an objective evaluation of behaviour is not present in LfD~\citep{argall2009survey}, where only expert demonstrations are available to be mimicked and generalised. 
The student can thus not surpass its master. 
Nevertheless, LfD is typically much more sample efficient than RL. 
Therefore, the aim is to combine the fast LfD method with objective behaviour evaluation and theoretical guarantees from RL. 

Two different approaches have been proposed to use demonstrations in an RL setting. 
The first is the generation of an initial value-function for temporal-difference learning by using the demonstrations as passive learning experiences for the RL agent~\citep{smart2002effective}. 
The second approach derives an initial policy from the demonstrations and uses that to kickstart the RL agent~\citep{brys2015reinforcement, suay2016learning}. 
In this regard, Taylor et al. propose the Human-Agent Transfer (HAT) algorithm~\citep{taylor2011integrating}, which consists of three steps: (i) demonstration: the agent performs the task teleoperated and records all state-action transitions, (ii) policy summarising: in order to bootstrap autonomous learning, policy rules are derived from the recorded state-action transitions, and (iii) independent learning: autonomous reinforcement learning using the policy summary to bias the learning.
Below we use the HAT algorithm to describe how RLfD fits into the ARL taxonomy.

\begin{itemize}
\item \textbf{Information source}: 
An expert of the task (human or otherwise) can provide sample behaviour by demonstrating its execution of the task. 
Preferably these demonstrations are efficient and successful executions of the task.

\item \textbf{Temporality}: 
It uses planned assistance.
Demonstrations are recorded and given to the learning agent before it starts training.

\item \textbf{Advice interpretation}: 
The received demonstrations must be first transformed into the agent's perspective by encoding them as sequences of state-action pairs. 
These are then processed using a classifier, which serves as the LfD component, creating an approximation of the demonstrator's policy using rules. 

\begin{figure}[t!]
\centering
\includegraphics[width=1\linewidth]{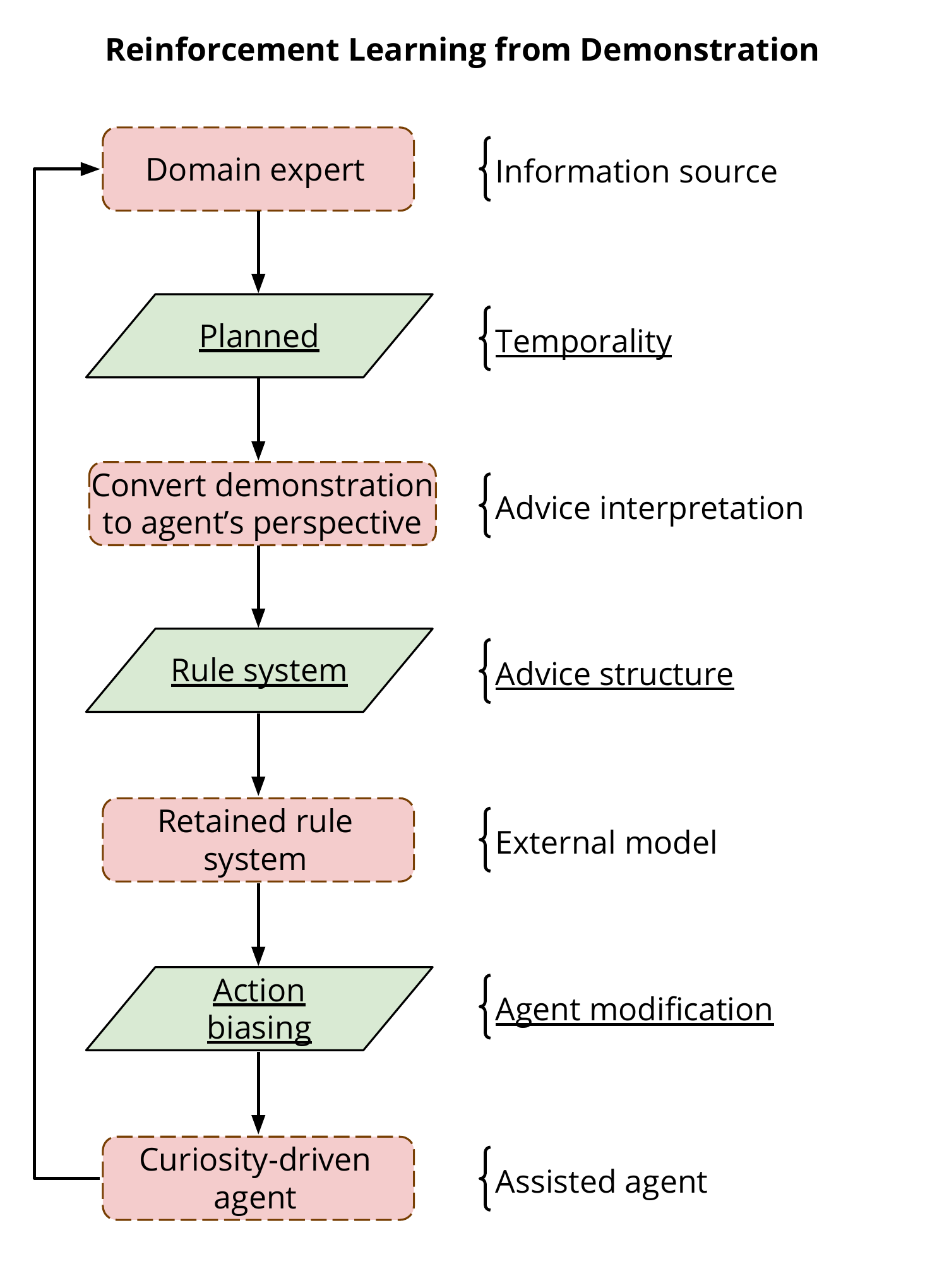}
\caption{RL from demonstration as the proposed ARL taxonomy. In this case, the processing components and communication links are defined from the HAT algorithm~\citep{taylor2011integrating}, which combines RL and LfD.}
\label{fig:ARL_Reinforcement Learning from Demonstration}
\end{figure}

\item \textbf{Advice structure}: 
The information is encoded as a classifier that maps states to the actions which the demonstrator is hypothesised to execute in those states.

\item \textbf{External model}: 
The generated rules are stored in the external model and not modified anymore. 
The external model can be queried with a state and responds with the hypothesised demonstrator action in that state.

\item \textbf{Agent modification}: 
The action proposed by the demonstrator can be integrated into the agent through three action biasing  methods: (i) attributing a value bonus to the Q-value for that state-action pair, (ii) extending the agent's action set with an action that executes the hypothesised demonstrator action, and (iii) probabilistically choosing to execute the action suggested by the model.

\item \textbf{Assisted agent}: 
During its decision-making (when and how depends on the implemented modification method) the agent has the option to consult the external model to obtain the action that the demonstrator is assumed to take.
This kind of agent is sometimes referred to as curiosity-driven agent~\citep{pathak2017curiosity}. 
Otherwise, the agent acts as a usual RL agent.
\end{itemize}

Figure~\ref{fig:ARL_Reinforcement Learning from Demonstration} shows how the RLfD approach is adapted to the proposed ARL taxonomy taking into account the previous definitions of processing components and communication links for the HAT algorithm.

\subsection{Transfer Learning}

\begin{figure}[t!]
\centering
\includegraphics[width=1\linewidth]{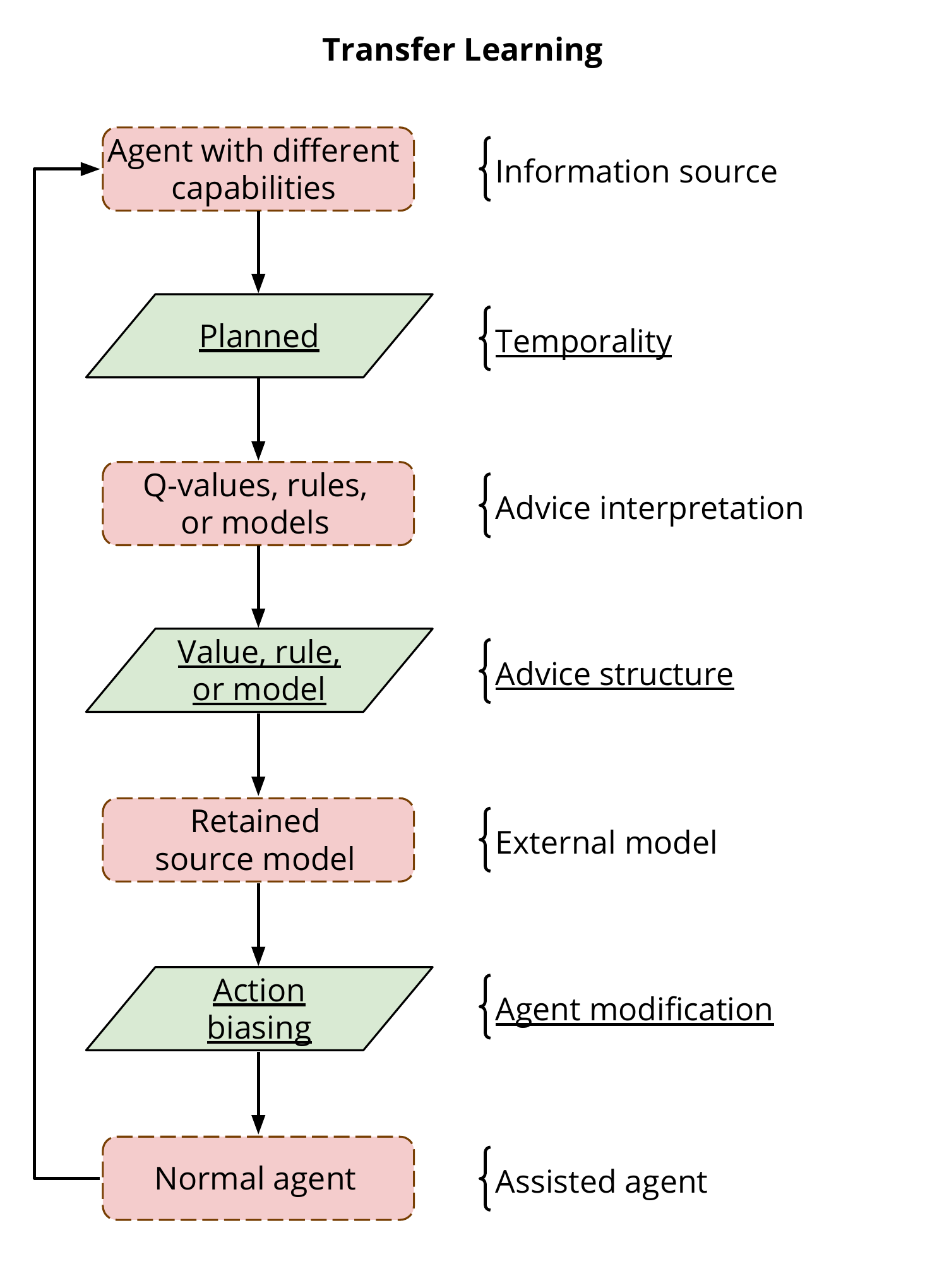}
\caption{Transfer learning as the proposed ARL taxonomy. In this case, an agent with different capabilities (or the same agent) provides the model of a source task which is transferred to a target task.}
\label{fig:ARL_Transfer Learning}
\end{figure}

The idea of transferring information between tasks (or between agents), rather than learning every task from the ground up seems to be obvious in retrospect. 
While transfer between different tasks has long been
studied in humans, it has only gained popularity in RL settings in the last decade~\citep{taylor2009transfer}. 
We consider three distinct settings where TL can be useful.

First, an agent may have learned how to perform a task and a new agent must learn to perform that same task or a variation on the task under different circumstances.
Let us consider two agents with different state features, i.e., different sensors, or different action spaces (or different actuators).
In this case, an inter-task mapping~\citep{ammar2011reinforcement, taylor2007transferJMLR} can be hand specified or learned from data~\citep{taylor2007transferAAMAS, ammar2015unsupervised} to relate the new target agent to the existing source agent. 
One of the simplest ways to reuse such knowledge is to embed it into the target task agent, e.g., directly reuse the Q-values that the source agent had learned~\citep{taylor2007transferJMLR}. 


Second, let us now consider that the world may be non-stationary. 
In TL settings, it is common to assume that the agent is notified when the world (or task in that world) changes.
However, a TL agent sometimes does not need to detect changes~\citep{hernandez-leal2016efficiently} or worry about the slow world changes over time~\citep{akila2015brief}. 
As in the previous setting, the agent may want to modify the information, e.g., by using an inter-task mapping, to relate the two tasks. 
In addition, the agent may decide not to use its prior knowledge at all, e.g., to avoid negative transfer because the tasks are too dissimilar~\citep{taylor2007transferJMLR}.

Third, TL could be a critical step within a curriculum learning approach~\citep{taylor2009assisting, bengio2009curriculum}. 
For example, previous work has shown that learning a sequence of tasks that gradually increase in difficulty can be faster than directly training on the final (difficult) task~\citep{taylor2007transferAAMAS, eppe2019curriculum}. 
In addition to curricula that are created by machine learning experts, curricula constructed by naive human participants have also been considered~\citep{peng2017curriculum}. 
Others have considered as a complementary problem a learning agent autonomously creating a curriculum~\citep{narvekar2017autonomous, dasilva2018object}. 
In all cases, the difficulty is scaffolding correctly so that the agent can learn quickly on a sequence of tasks. 
These approaches are distinct from multi-task learning~\citep{fernandez2006probabilistic}, where the agent wants to learn over a distribution of tasks, and lifelong learning~\citep{chen2016lifelong, parisi2019continual}, where learning a new task should also improve performance on previous tasks.
The following is an analysis of TL methods in terms of the ARL taxonomy.

\begin{itemize}
\item \textbf{Information source}: 
The information comes from an agent with different capabilities or the same agent that has trained on a different task.

\item \textbf{Temporality}: 
Transfer typically occurs when a task changes or when an agent first faces a novel task. 
In both cases, it is planned assistance, i.e., the source agent transfers knowledge to the target agent before the target agent begins learning. 
If the inter-task mapping is initially unknown, some time may be spent trying to learn an inter-task mapping or estimate task similarity to previous tasks. 
However, the more time spent before the transfer, the less impact transfer can have.

\item \textbf{Advice interpretation}:  
There are many types of information that can be transferred, including Q-values, rules, a model, etc.~\citep{taylor2007transferJMLR}. 
TL methods assume the target agent has access to the source agent's `brain', an assumption that may not always be true, e.g., if the designer of the source agent has not provided an API or if the source agent is a human.

\item \textbf{Advice structure}: 
The structure of the transferred knowledge is as varied as the types of information that can be provided.
This variety of information includes Q-values, rules, or a model, among others.

\item \textbf{External model}: 
The source model is normally retained.
Because the source task knowledge is not necessarily sufficient for optimal performance in the target task, it is important for the target agent to be able to learn to outperform the transferred information.

\item \textbf{Agent modification}: 
The target task agent uses the transferred information to bias its learning. 
The transferred knowledge is not typically modified. 
Instead, the target task agent builds on top of the knowledge, learning when to ignore it and instead follow the knowledge it has learned from the environment.

\item \textbf{Assisted Agent}: 
The agent is a typical RL agent that can take advantage of one or more types of prior knowledge. 

\end{itemize}

Figure~\ref{fig:ARL_Transfer Learning} shows how the TL approach can be represented within the proposed ARL taxonomy taking into account the previous definitions of processing components and communication links.

\subsection{Multiple Information Sources}
\label{subsec:multiple_information_sources}

While the majority of work in ARL is based on a single source of advice, several researchers have considered scenarios where multiple sources of advice may exist~\citep{brys2017multi, dasilva2019integrating, gimelfarb2018reinforcement, yamagata2019online}. 
Although the use of multiple information sources is not an ARL approach by itself and could comprise sources utilising any of the previously mentioned approaches, we include it here to highlight how this multiple sources can be framed within the proposed taxonomy. 
The introduction of multiple advisors may have benefits for ARL agents, particularly in scenarios where each individual advisor has knowledge which is limited in some way~\citep{shelton2001balancing}, e.g., individual advisors may have expertise covering different sub-areas of the problem domain. 
However, it also introduces additional problems for the agent, such as handling inconsistencies or direct conflicts between the guidance provided by different advisors, or learning to judge the reliability of each advisor, possibly in a state-sensitive manner~\citep{zhan206theoretically}. 
In the extreme case, an agent may even need to be able to identify and ignore the advice provided by deliberately malicious advisors~\citep{nunes2003exchanging}.
The following is an analysis of approaches using multiple information sources with respect to the proposed ARL taxonomy.

\begin{itemize}
\item \textbf{Information source}: 
Prior research has identified several scenarios in which an agent may have access to multiple sources of external information. 
Argall et al.~\citep{argall2009automatic} argue that when robots are applied to tasks within society in general, it is very likely that multiple users will interact with and guide the behaviour of a robot. 
In the context of TL, multiple sources of information may be derived either from experience on varying MDPs~\citep{parisotto2015actor}, or on alternative mappings from a single prior MDP to the current environment~\citep{talvitie2007experts}. 
In multi-agent systems, each agent may serve as a potential source of information for every other agent~\citep{dasilva2017simultaneously, fachantidis2019learning}.

\item \textbf{Temporality}: 
Assistance may be planned or interactive. 
For instance, Argall et al.~\citep{argall2009automatic} have considered two different sources of information, in the form of teacher demonstrations and teacher feedback on trajectories generated by the learner. 
The former may be provided in advance of learning consisting of complete state-action trajectories, i.e., planned assistance, while the latter occurs on an interactive basis during learning, and structurally consists of a subset of the learner's actions being flagged as correct by the teacher, i.e., interactive assistance.

\item \textbf{Advice interpretation}: 
The majority of work so far on ARL from multiple information sources has assumed that these sources are homogeneous in terms of the timing and nature of the information provided. 
However, this need not be the case, and for heterogeneous information sources, some aspects of the advice may differ in terms of interpretation and structure.
In this regard, the advice needs to be integrated considering either all possible sources (equally or non-equally contributing), some sources (with the information provided partially or fully considered), or only from one source at a time~\citep{shelton2001balancing}.

\item \textbf{Advice structure}: 
Each information source may use a different structure of advice.
Therefore, individually all the aforementioned structures in previous sections are possible to be used, e.g., machine rule, state-action pair, rule system, value, or model.
The final structure into a single piece of advice may be done by integrating the multiple information sources, for instance using a multi-modal integration function~\citep{cruz2016multi} or using graph structures (e.g., graph neural networks) using causal links between features for multi-modal causability~\citep{holzinger2021towards}.

\begin{figure}[t!]
\centering
\includegraphics[width=1\linewidth]{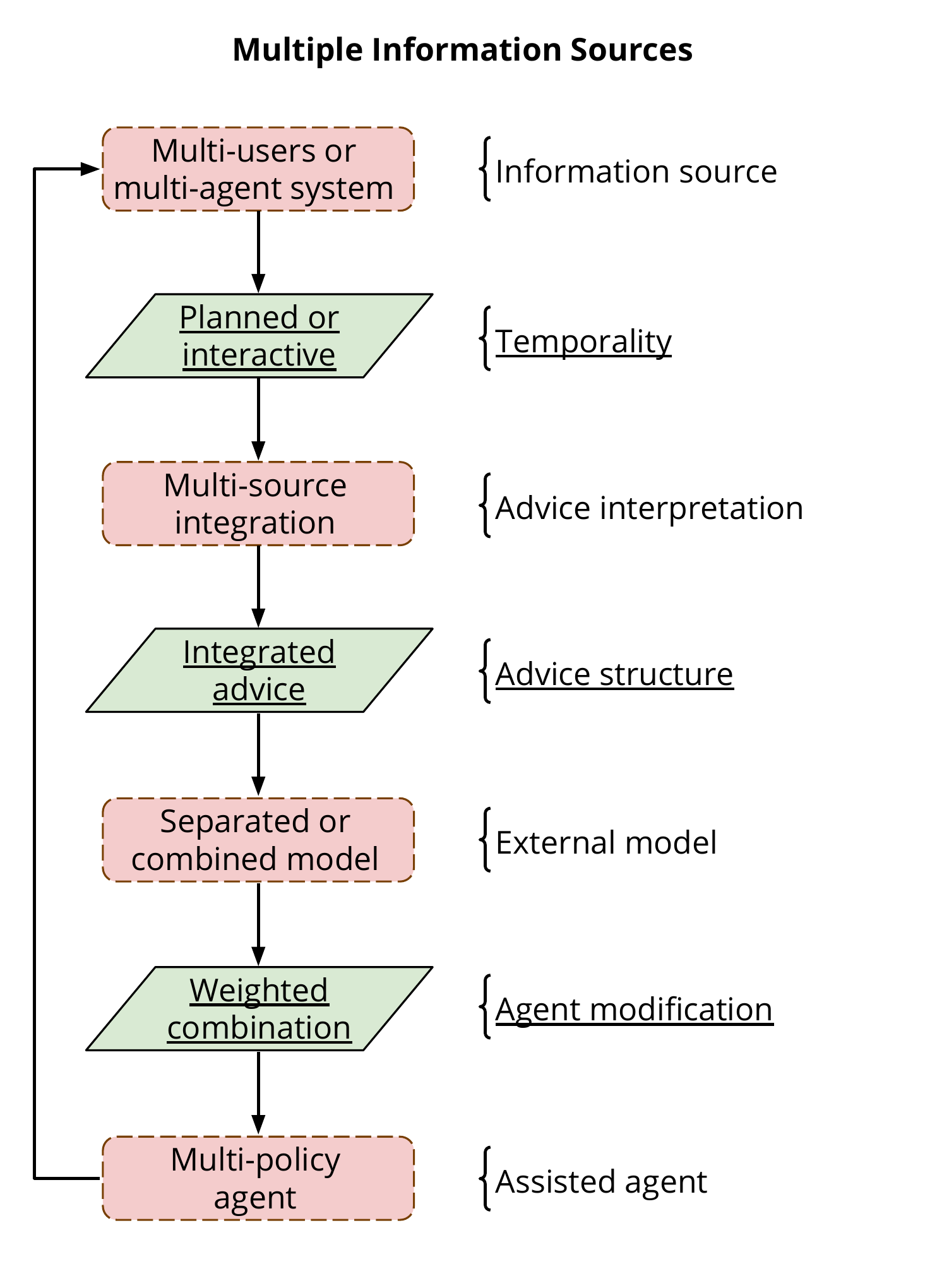}
\caption{Multiple information sources as the proposed ARL taxonomy. In this case, there could be multiple humans or multiple agents. One important aspect is to integrate the different pieces of advice. The agent may also learn multiple policies as in multi-objective RL.}
\label{fig:ARL_Multiple Information Sources}
\end{figure}

\item \textbf{External model}: 
An ARL agent must choose whether (i) to maintain a separate model for each information source, (ii) to combine the information from all sources into a single model, or (iii) a combination of both. 
An example of the latter approach is the inverse RL system presented in~\citep{karlsson2014learning}, which learns a model of each information source in the form of a feature-weighting function and then forms a combined feature-weighting via averaging. 
As noted by Karlsson~\citep{karlsson2014learning}, single-model approaches may encounter difficulties if dealing with information sources which are fundamentally incompatible with each other. 
An additional benefit of maintaining independent models is that these can also be augmented by additional data on characteristics of each information source, such as the reliability or consistency of its advice~\citep{argall2009automatic, talvitie2007experts}. 

\item \textbf{Agent modification}: 
Any of the modification approaches discussed in the earlier sections of this paper may also be applied in the context of multiple information sources.
For example, agent modification methods from LfD~\citep{argall2009automatic}, TL~\citep{talvitie2007experts, parisotto2015actor}, reward-shaping~\citep{brys2014combining, knox2013training} as well as inverse RL~\citep{karlsson2014learning, tanwani2013transfer}. 
The main additional consideration is how these methods may be affected by the presence of multiple external models. 
The main methods examined so far use a combination of the models, either weighted or unweighted~\citep{argall2009automatic, karlsson2014learning} or select a single best model to use~\citep{talvitie2007experts}.

\item \textbf{Assisted Agent}: 
In most circumstances, the operation of the agent itself is largely unaffected by the presence of more than one information source. 
However, Tanwani and Billard~\citep{tanwani2013transfer} consider the task of performing inverse RL from multiple demonstrations provided by multiple experts, operating according to different strategies or preferences. 
To address the potential incompatibilities between these strategies, the agent attempts to learn a set of multiple policies, so as to be able to satisfy any policy expert strategy, including those not provided to the agent. 
This approach is closely related to multi-policy algorithms developed for multiobjective RL~\citep{roijers2013survey}. 
\end{itemize}

Figure~\ref{fig:ARL_Multiple Information Sources} shows how an approach using multiple information sources is adapted to the proposed ARL taxonomy taking into account the previous definitions of processing components and communication links.
Moreover, Table~\ref{tab:ARLSummary} summarises how each of the ARL approaches and examples reviewed in this section is adapted to the proposed taxonomy.

\begin{table*}[t!]
\centering
\caption{Summary of the reviewed assisted reinforcement learning approaches adapted to the proposed taxonomy.}

\begin{tabular}{| L{2.2cm} | L{1.8cm} | L{1.6cm} | L{1.6cm} | L{1.6cm} | L{1.7cm} | L{1.8cm} | L{1.6cm} |}
\hline
\textbf{\thead{Approach}} & \textbf{\thead{Information\\source}} & \textbf{\thead{Tempora-\\lity}} & \textbf{\thead{Advice\\interpre-\\tation}} & \textbf{\thead{Advice\\structure}} & \textbf{\thead{External\\model}} & \textbf{\thead{Agent\\modifi-\\cation}} & \textbf{\thead{Assisted\\agent}} \\    \hline    \hline
Heuristic reinforcement learning & Human-domain expert & Planned & Convert rule to machine language & Machine rule & Retained rule-set & Policy shaping & Normal agent\\[5ex]
\hline
Interactive reinforcement learning & Human / simulated user & Interactive & Convert modal cue to signal & State-action pair & Immediate & Policy / reward shaping & Curiosity-driven agent\\[5ex]
\hline
Reinforcement learning from demonstration & Domain expert & Planned & Convert demonstration to agent's perspective & Rule system & Retained rule system & Action biasing & Curiosity-driven agent\\[5ex]
\hline
Transfer learning & Agent with different capabilities & Planned & Q-values, rules, or models & Value, rule, or model & Retained source model & Action biasing & Normal agent\\[5ex]
\hline
Multiple information sources & Multi-users or multi-agent system & Planned or interactive & Multi-source integration & Integrated advice & Separated or combined model & Weighted or unweighted combination & Multi-policy agent\\[5ex]
\hline
\end{tabular}
\label{tab:ARLSummary}
\end{table*}

\section{Future Directions and Open Challenges}
\label{sec:future_work}

In this section, we discuss open issues and propose further possibilities for future work in the field of ARL. 
These open questions have been identified from the current literature in the field. 
Many of these issues are shared with autonomous RL but it still remains open how they could be addressed within the ARL framework.

\subsection{Incorrect Assistance}
\label{subsec:incorrect_advice}
A common assumption that ARL methods make is that all external information that the agent receives is accurate~\citep{efthymiadis2013overcoming}. 
Accurate information is correct advice that assists the agent in completing its goal. 
However, the assumption that information will always be of use to the agent is wrong, especially when the information source is an observing human, as in RL from imperfect demonstrations~\citep{gao2018reinforcement, jing2020reinforcement}. 
Humans may deliver advice late, and therefore the agent may relate it to a wrong state.
The advice may be of short-term use to the agent but prevent it from achieving optimal performance.
Moreover, the human trainer may even be malicious and actively attempting to sabotage the agent's performance. 

Incorrect information can be introduced by other sources as well. 
Some examples for non-human incorrect advice include behaviour transferred from another domain that does not align correctly, rules that generalise over multiple states which may cover exception states, and noisy or missing information from audio-visual sources~\citep{cruz2016multi}.

Information given to agents may be correct initially, but over time no longer be the optimal solution~\citep{akila2015brief}. 
Other advice may be mostly accurate or correct for most states, however, there can exist states of exception to the advice. 
These exception states can be the critical difference between an ordinary solution and the optimal solution. 
There is a need for research on how to identify and mitigate incorrect information in these scenarios, especially considering that even a very small amount of incorrect advice may be really detrimental for the learning process~\citep{cruz2018improving}.

\subsection{Multiple Information Sources}
\label{subsec:multiple_information_sources_question}

As reviewed in the previous section, the use of multiple information sources may naturally arise on some application scenarios, and can increase the agent's knowledge of the environment, and increase confidence in decision-making if the different sources agree on an action. 
However, the use of multiple sources raises additional questions: 

\begin{itemize}
\item What if the different sources disagree on the best action to take?
\item How can the agent identify the best information source to listen to?
\item How can the agent manage conflicting information?
\item How can the agent measure trust in the different information sources?
\end{itemize}

Additionally, the use of multiple sources may be extended to crowdsourcing~\citep{kamar2012combining}. 
In this context, crowdsourcing refers to the enlistment and use of a large number of people, either paid or unpaid and can range in size from tens to tens of thousands. 
Typically, crowdsourcing is performed via the internet. 
This can raise challenges of malicious users, anonymity, and large uncertainty in the value and reliability of the information. 


\subsection{Explainability}
\label{subsec:interpretability}
Explainability refers to translating the agent's information into a form the human can understand~\citep{cruz2019memory, dazeley2021levels}.
The reasons why an agent develops certain behaviours can sometimes be difficult to understand for non-expert end-users. 
Systems to measure the quality of explanations generated by AI-based systems have been previously introduced in order to build effective and efficient human-AI interaction~\citep{holzinger2020measuring}. 
When combining the RL method with policy modification methods such as rules, expert assistance, external models, and policy-shaping, understanding why an agent chooses to take an action becomes even more difficult. 
Developing methods for understanding agent learning and its decision-making is important as it allows the human to remain informed of the agent's motivations and decisions, and keep track of the accountability of the actions taken~\citep{dazeley2021conceptual}. 
This can be beneficial for artificial intelligence ethics, and human-computer teaching, among other fields. 

\subsection{Two-Way Communication}
\label{subsec:two_way_communication}
Two-way communication refers to the ability for the information source and the agent to converse with each other, perhaps multiple times before making a decision~\citep{kessler2019active}.
Two-way communication can allow the information source, presumably human, and the agent to ask questions to each other, request more information, and to clarify decision-making and its reasoning. 
Although the proposed framework includes two-way communication, as shown in Figure~\ref{fig:ARL_Intro}, most current ARL methods do not have two-way communication to the extent that non-expert human advisors can interact with the agent freely. 
For two-way communication to apply to non-expert human advisors issues of explainability (as shown in the previous section), timing, and agent initiation need to be addressed. 

Timing refers to the time it takes to communicate back and forth. 
Agents sometimes have a fixed time limit, during which they need to learn, communicate, and decide on the next action. 
Methods for reducing the time it takes to interact with the human and reducing the number of interactions needed with the human are two areas open for research.
Agent initiation refers to the ability for the agent to initiate communication with the human source itself. 
The agent may choose to do this so to request clarification on information, or request assistance for decision-making. 
A challenge for agent initiation is to determine when and how often the agent should request assistance. 
The requests for assistance should be frequent enough to make use of the information source while not becoming a nuisance to the human, or detracting from learning time, and should consider the cost of the request, e.g., in paid crowdsourcing.

\subsection{Other Challenges}
There are also other challenges to be considered for future possibilities of ARL systems.
Although many of the issues described in this section are also shared with autonomous RL~\citep{mankowitz2019challenges}, we focus the discussion on how particularly externally-influenced agents may be affected in the context of the ARL framework.
While we describe the essential implications on ARL systems for each of the following areas, we note that further and deeper discussion may be addressed for each of them.

\begin{itemize}
    \item \textbf{Real-time policy inference}: Many RL systems need to be deployed in real-world scenarios and, therefore, policy inference must happen in real-time \citep{koenig1993complexity}.
    Using ARL frameworks may lead to additional issues since the external information source should observe and react to the RL agent's state as fast as possible, otherwise the assistance may become unnecessary or incorrect for the new reached state.

    \item \textbf{Assistance delay}: There are RL systems where determining the state or receiving the reward signal may take even weeks, such as a recommender system where the reward is based on user interaction~\citep{mann2018learning}. 
    In these contexts, the external information source may also lead to unknown delays in the system actuators, sensors, or rewards, making the assistance atemporal, either delayed or ahead, or even in some cases being conflicting or redundant considering the RL agent's autonomous operation.
    
    \item \textbf{Continuous states and actions}: When an RL agent works in high-dimensional continuous state and action spaces~\citep{millan2019human, ayala2019reinforcement} there could be issues for learning even in traditional RL~\citep{dulac2015deep}. 
    In an ARL framework, additional problems may be present as the agent uses external information which may be not accurate enough given the high dimensionality. 
    In the presence of high-dimensional states and actions, even small differences in the received assistance may substantially slow the learning process since these differences may represent in essence a very different state or action.
    
    \item \textbf{Safety constraints}: In RL environments, there are safety constraints that should never or at least rarely be violated~\citep{karimpanal2019learning}.
    Special care is needed when receiving information from an external source since there could be situations that the advisor may repeatedly direct the agent to unsafe states and, in turn, lead to an increase in the time needed for learning.
    
    \item \textbf{Partially observable environments}: In practice, many RL problems are partially observable~\citep{chen2018partially}.
    For instance, partial observabilities may occur in non-stationary environments~\citep{millan2019human} or in presence of stochastic transitions~\citep{cruz2021explainable}.
    If the external information source does not have observations to clearly infer the current state in the environment may lead to giving incorrect assistance to the learner agent. 

    \item \textbf{Multi-objective reward}: In many cases, RL agents need to balance multiple and conflicting subgoals, therefore, they may use multi-dimensional reward functions~\citep{vamplew2020demonstration}.
    In this regard, an external information source may give priority to a particular subgoal over the others, unbalancing the global reward function.
    There could be also issues when multiple information sources are used covering or favouring different subgoals.
    Moreover, when using a multi-objective reward in TL, there could only be some subgoals from the source task which are relevant in the target task, therefore, the RL agent should also coordinate and filter relevant information.
    
    \item \textbf{Multi-agent systems}: There could be multiple agents learning a task and multiple external information sources.
    In this case, if an information source provides advice it could be generalised to all of them or it could be pointed specifically to an agent.
    Moreover, advice useful for one agent may be detrimental to another, depending on the state, the agent's current knowledge, or its particular reward function. 
    Using multiple information sources, if an agent consults an external source, it may be necessary to discriminate which one is the best for the particular state.
    Additionally, the teacher-student approach usually integrated into ARL requires the teacher to be an expert in the learning domain.
    In this regard, multiple learning agents may also advise each other while learning in a common environment~\citep{dasilva2017simultaneously}.    
\end{itemize}

\section{Conclusions}
\label{sec:conclusion}
In this article, we have reviewed ARL methods and presented an ARL framework, comprising all RL techniques that use external information. 
ARL methods use external information to supplement the information the agent receives from the environment to improve performance and decision-making. 

To describe the different ARL methods, we propose a taxonomy to classify the different functions of an externally-influenced RL agent. 
Through the analysis of the current literature, we have found seven key features that make up an ARL technique. 
They are divided into four processing components and three communication links.
A definition and examples of each of these seven features have been presented.

The contribution of this paper is twofold: the review of state-of-the-art ARL methods and the ARL taxonomy as an additional level of abstraction. 
However, future work framed into our proposed ARL taxonomy can also make use of the different concepts here defined, either processing components or communication links. 
In this regard, it is essential to understand that not each ARL method must necessarily use all the proposed concepts. In some cases, simplified models may also be a representation of the ARL framework.

Additionally, we demonstrated the applicability of the framework on different ARL fields. 
These areas include heuristic RL, IntRL, RLfD, TL, and multiple information sources. 
Each of these fields has been analysed and described as applied to the presented taxonomy. 
Finally, we also present some ideas about areas for future research in order to extend the ARL field.

\section*{Acknowledgments}
This work has been partially supported by the Australian Government Research Training Program (RTP) and the RTP Fee-Offset Scholarship through Federation University Australia.

\bibliographystyle{ieeetr}

\balance
\bibliography{paper}

\end{document}